\pdfoutput=1

\documentclass[11pt]{article}

\usepackage[]{emnlp2021}

\usepackage{times}
\usepackage{latexsym}

\usepackage{graphicx} 
\usepackage{float} 
\usepackage{subfigure} 
\usepackage[T1]{fontenc}

\usepackage[utf8]{inputenc}

\usepackage{microtype}

\title{Discovering Representation Sprachbund For Multilingual Pre-Training}

\author{Yimin Fan\textsuperscript{1}\thanks{\hspace{2mm}Work is done during  internship at Microsoft Research Asia.}, Yaobo Liang\textsuperscript{2}, Alexandre Muzio\textsuperscript{3}, Hany Hassan\textsuperscript{3}, \\ \bf{Houqiang Li\textsuperscript{1}}, \bf{Ming Zhou\textsuperscript{4}} and \bf{Nan Duan\textsuperscript{2}} \\
        \textsuperscript{1}University of Science and Technology of China, \textsuperscript{2}Microsoft Research Asia, \\ \textsuperscript{3}Microsoft, \textsuperscript{4}Sinovation Ventures \\
       \texttt{fym0503@mail.ustc.edu.cn}, \texttt{lihq@ustc.edu.cn} \\
       \texttt{\{yalia, nanduan, alex.muzio, hanyh\}@microsoft.com}, \\ \texttt{mingzhou926@hotmail.com}}

\begin{document}
\maketitle
\begin{abstract}
Multilingual pre-trained models have demonstrated their effectiveness in many multilingual NLP tasks and enabled zero-shot or few-shot transfer from high-resource languages to low-resource ones. However, due to significant typological differences  and  contradictions  between  some languages, such models usually perform poorly on many languages and cross-lingual settings, which  shows  the  difficulty  of  learning  a  single  model  to  handle  massive diverse languages  well  at  the same time. 
To alleviate this issue, we present a new multilingual pre-training pipeline. We propose to generate language representation from multilingual pre-trained models and conduct linguistic analysis to show that language representation similarity reflect linguistic similarity from multiple perspectives, including language family, geographical sprachbund, lexicostatistics and syntax.
Then we cluster all the target languages into multiple groups and name each group as a representation sprachbund.
Thus, languages in the same representation sprachbund are supposed to boost  each  other  in  both  pre-training and fine-tuning as they share rich linguistic similarity.
We pre-train one multilingual model for each representation sprachbund. Experiments  are  conducted  on  cross-lingual benchmarks and significant improvements are achieved compared to strong baselines.

\end{abstract}

\section{Introduction}
The use of pre-trained models is considered a milestone in the development of NLP research. Though early works~\citep{devlin-etal-2019-bert,radford2019language} on monolingual pre-training (pre-training one model for one language) significantly boosts the performance on the target language, monolingual pre-training can hardly be generalized to multilingual settings because of high training cost and insufficient corpora resources for many languages.

Multilingual pre-training was proposed to resolve this issue. By using shared vocabulary across languages and  pre-training with corpora from multiple languages, multilingual pre-trained models handle cross-lingual tasks in one model. Large scale multilingual pre-trained models provide powerful representation for languages worldwide, enabling significant advances in various multilingual tasks. However, existing widely used multilingual pre-trained models~\citep{DBLP:journals/corr/abs-1901-07291,conneau-etal-2020-unsupervised,huang-etal-2019-unicoder} perform poorly on many languages and some cross-lingual tasks like zero/few-shot cross-lingual transfer. \emph{e.g.} the performance of zero shot transfer on XNLI task from English data to Urdu language is 15\%+ lower than to English~\citep{conneau-etal-2020-unsupervised}. Such a huge performance gap is the result of cross-lingual contradictions and differences. This phenomenon is also recognized as negative transfer in transfer learning~\citep{Wang_2019_CVPR}.

Many existing works in cross-lingual transfer~\citep{K2020Cross-Lingual,pires-etal-2019-multilingual,lin-etal-2019-choosing} and machine translation~\citep{dabre-etal-2017-empirical,tan2019multilingual} have shown that cross-lingual transfer works best between typologically similar languages. We believe that utilizing similarity between languages is potentially beneficial for large-scale multilingual pre-training. Motivated by this, we propose a new multilingual pre-training pipeline. First, we design a fully data-driven end-to-end way to generate language 
representation for all languages (108 languages) based on massive multilingual corpora. We represent each language as a 768-dimension vector and use cosine similarity as their similarity measure. With the similarities between languages quantified by their language representation, we automatically divide all languages into a small number of representation sprachbunds. Sprachbund is a linguistic terminology in German that refers to a group of close languages (Sprach means language and bund means federation in German, so literally it is ``language federation" in English). A representation sprachbund is defined as a group of languages with similar language representation. We conduct extensive linguistic analysis and show that languages with similar representation are similar and related from many linguistic perspectives, including language family, geographical sprachbund, lexicostatistics and syntax typology. We believe that training with similar languages in pre-training and fine-tuning are beneficial 
as they share similar linguistic properties.
Second, we train multiple multilingual pre-trained models. Each model is trained with corpora from one representation sprachbund. When handling downstream tasks in one specific language, we fine-tune the model pre-trained with the corresponding representation sprachbund corpora. We conduct experiments on 8 representative cross-lingual tasks from XGLUE~\citep{liang-etal-2020-xglue} and XTREME~\citep{hu2020xtreme} including sentence classification, structure prediction, question answering and sentence retrieval. Experiment results show that our model significantly outperforms strong baselines.

Our contributions can be summarized as follows: i) We propose a way to automatically generate language representation from multilingual pre-trained models and massive multilingual corpora.  ii) We conduct extensive  analysis to show language representation and representation sprachbunds can reflect linguistic language similarity and relatedness from multiple perspectives, therefore they can be considered as new paradigm for clustering similar languages in linguistics.  iii) We use representation sprachbunds in multilingual pre-training to alleviate the cross-lingual contradiction and differences, and obtain significant improvements compared with strong baselines.

\section{Related Work}
Our approach presents a new pipeline of multilingual pre-training. Our representation sprachbund is inspired by linguistic language clustering. Our analysis is closely related to methodology in linguistics. 
\paragraph{Multilingual Pre-Training}
Multilingual pre-training was proposed to pre-train a single model with hundreds of languages. Many works use a large amount of multilingual data (\emph{e.g.}, mC4~\citep{xue2020mt5}, CCNet~\citep{wenzek2020ccnet}) to pre-train large multilingual models like XLM~\citep{DBLP:journals/corr/abs-1901-07291}, XLM-R~\citep{conneau-etal-2020-unsupervised}, Unicoder~\citep{huang-etal-2019-unicoder} and mT5~\citep{xue2020mt5}. Several benchmarks are proposed to evaluate the cross-lingual ability of multilingual pre-trained models, including XGLUE~\citep{liang-etal-2020-xglue}, XTREME~\citep{hu2020xtreme} and XTREME-R~\citep{ruder2021xtremer}

\begin{figure*} 
\centering 
\includegraphics[width=0.9\textwidth]{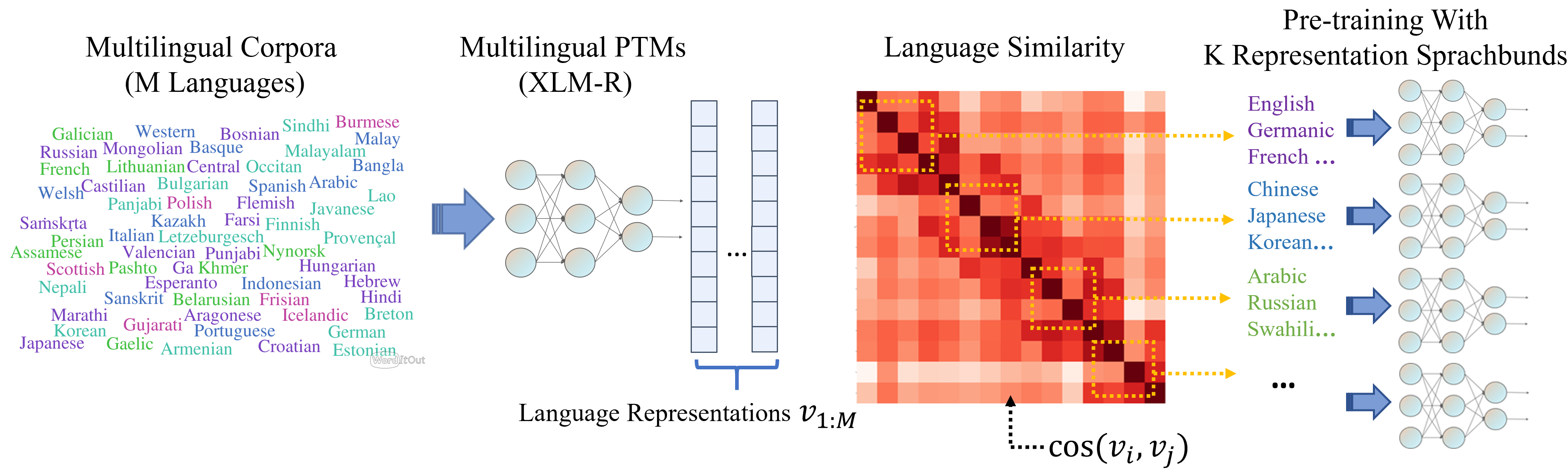} 
\caption{The pipeline of our approach. We first generate language representation for each language with multilingual pre-trained models (XLM-R) and multilingual corpora. We cluster languages into several representation sprachbunds composed of languages with similar representation. We pre-train one model for each representation sprachbund with corpora from that representation sprachbund. Best viewed in color.  
} 
\label{Fig.main2} 
\end{figure*}
\paragraph{Language Clustering in Linguistics} The linguists propose to classify languages in several ways from different perspectives. There are two main kinds of language clustering: genealogical clustering and typological clustering. In genealogical clustering, languages are clustered into language families~\citep{durbin1985survey,marcantonio2002uralic} by their genetic relatedness. Languages in the same language family have the same ancestral language. In typological clustering, languages are clustered by their typological features, like word order, morphology~\citep{DRESSLER1986197} and lexicostatistics~\citep{10.2307/2739673}. Geographical sprachbund~\citep{nla.cat-vn19682} is also a typological clustering method as it groups languages according to their similar areal features coming from geographical proximity. 
\paragraph{Language Clustering in Multilingual NLP}
Several recent works utilize linguistic knowledge about language clustering in multilingual pre-training and machine translation. \citet{tan2019multilingual} uses language family and language embedding to cluster languages and train machine translation model for each cluster. The language embedding is the language-specific tag added to the input of encoder. Their approach focuses on 23 relatively high resource languages. \citet{fan2020englishcentric} clusters languages into several groups according to language family, cultural connection and geographical proximity. They do not obtain any language representation and their language groups are human annotated. \citet{kudugunta-etal-2019-investigating} reveals the connection between language SVCCA similarity from NMT models and language family. Their evaluation relies on parallel data. \citet{chung2020improving} classifies languages into groups based on their token overlap. Only lexical information of languages is used in their approach. \citet{yu-etal-2021-language} uses multilingual denoising autoencoder to generate language embeddings and analyze the clusters derived from the embeddings. There are also a few earlier works on generating and analyzing language representation \cite{tiedemann2018emerging,ostling-tiedemann-2017-continuous}.

Compared to existing works, our approach enjoys the following advantages. First, our approach requires neither parallel data nor linguistic labeling, while most existing works on clustering languages relies on parallel data and linguistic knowledge. Second, our representation sprachbunds contain linguistic features from various aspects, while most existing works focus on language family.

\section{Approach}
 Our proposed pipeline on representation sprachbund for multilingual pre-training is illustrated in Figure~\ref{Fig.main2}. Our approach can be divided into two stages: First, we quantify the similarity between languages with generated language representation, and cluster all languages in our corpora into $K$ clusters based on their similarity. Each cluster is called one representation sprachbund. $K$ is a hyper-parameter in our clustering algorithm.
 Second, we separate our corpora into $K$ parts based on their corresponding representation sprachbund, and pre-train one model with corpora from each representation sprachbund.
 \subsection{Discovering Representation Sprachbund}
\label{section3.1}
Suppose we have $M$ languages in our multilingual corpora $C$ denoted as $L=\{l_1,l_2,...,l_M\}$. The corpora of the $i$th language is denoted as $C_i$. $C_i$ contains $n_i$ sentences, denoted as $C_i=\{s_{i1},s_{i2},...,s_{in_i}\}$. Note that there is no need for the sentences $\{s_{ik}\}_{k=1}^{n_i}$ of language $l_i$ and $\{s_{jk}\}_{k=1}^{n_j}$ of $l_j$ to be aligned.

\citet{choenni2020does} reveals that the sentence representation from the same language generated by the last layer of multilingual pre-trained models will be very close and form a relatively independent cluster in the representation space. Motivated by this, a centroid of all sentence representation from the same language can be a reasonable language representation.
We employ a transformer-based multilingual pre-trained model, denoted as $F$. Each language $l_i$ is represented by language representation $v_i$. We denote the representation of the token $k$ of sentence $s$ from the last layer of $F$ as $F_k(s)$. We then define 
$$
v_i=\frac{1}{n_i}\sum_{j=1}^{n_i}F_{[CLS]}(s_{ij})
$$

We separate all $M$ languages into $K$ clusters via clustering algorithm with input features $\{v_i\}_{i=1}^M$. The cosine similarity between language representation $v_i$ and $v_j$ is used as a similarity metric for the clustering algorithm.
The output $K$ clusters are $K$ representation sprachbunds.

\begin{figure*} 
\centering 
\includegraphics[width=0.9\textwidth]{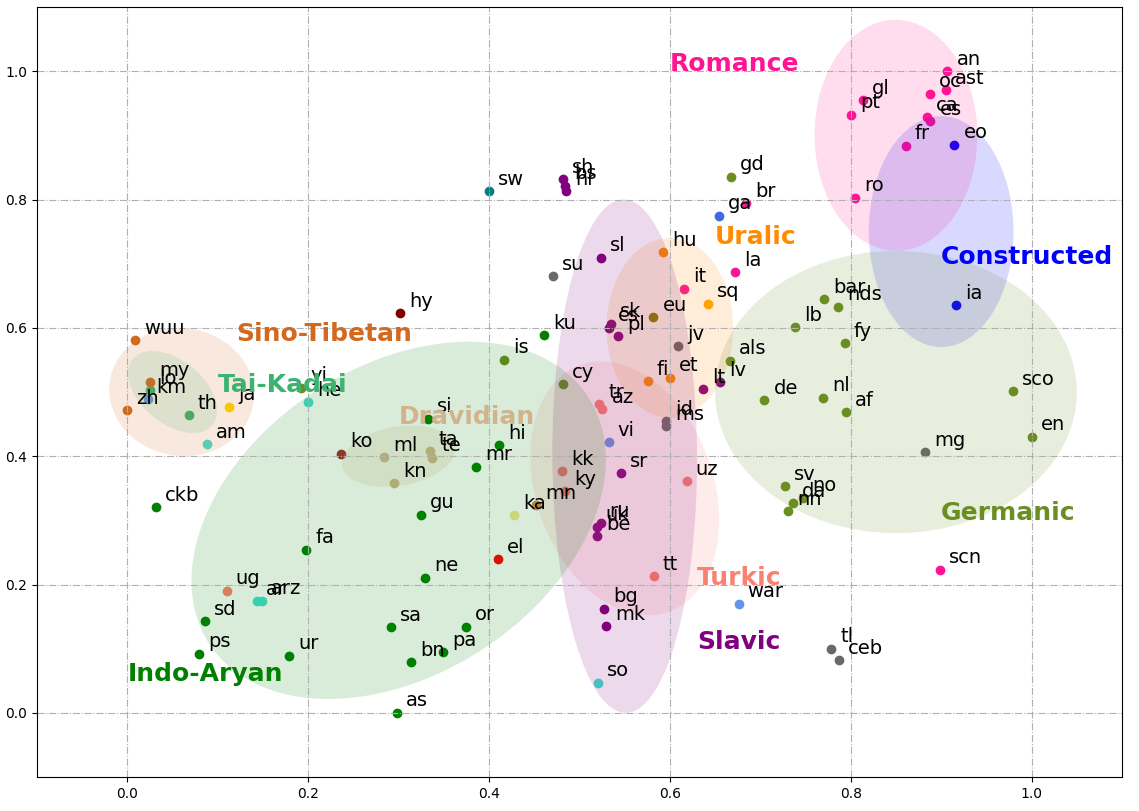} 
\caption{Visualization of language representation (reduced to 2-dimension for visualization). All languages are labeled with ISO 639-1 code. Languages from the same language family are colored the same. We draw ellipse for 10 main language family covering most languages they include. The distribution of language representation has great overlap with the language family and several geographical sprachbunds. Best viewed in color. } 
\label{Fig.main_vis} 
\end{figure*}
\subsection{Representation Sprachbund for Multilingual Pre-Training}

We denote our $K$ representation sprachbunds as $\{L_1,L_2,...,L_K\}$, where $L_k=\{l_{k1},...,l_{kk_i}\}$. We have $\sum_{j=1}^Kj_i=M$ as our representation sprachbund is non-overlapping. The corresponding corpora is denoted as $C'_k=\{C'_{k1},C'_{k2},...,C'_{kk_i}\}$. Note that the corpora of the $i$th language for discovering representation sprachbund ($C_i$) and for training models ($C'_i$) may be different.
We train $K$ separate models for $K$ representation sprachbunds. When fine-tuning, we can use the data in language $\{l_{kj}\}_{j=1}^{k_i}$ to fine-tune the $k$th model if the data is available.


\section{Representation Sprachbund Discovery and Analysis}
\subsection{Settings}
\label{RSBst}
We collect massive multilingual corpora for discovering representation sprachbund (also for the following multilingual pre-training). We use Wikipedia\footnote{\url{https://en.wikipedia.org/wiki/Main_Page}} corpora (100 languages are included, total size 101GB) and a clean version of Common Crawl (CC)\footnote{\url{https://commoncrawl.org/}} (89 languages are included, total size 2500GB) following \citet{liang-etal-2020-xglue}. 108 languages are included in our multilingual corpora. Note that we do not use any parallel corpora. The pre-trained model $F$ we use is XLM-R base model implemented by HuggingFace\footnote{\url{https://huggingface.co/xlm-roberta-base}}. 
As the size of the whole corpora is very large, we use a random sampling strategy to get part of the data for extracting representation. For those languages with less than 10GB of data, we use all the data for extracting representation; for those languages with more than 10GB of data, we sample 10GB out of all the data. We use the method mentioned in Section~\ref{section3.1} to get the language representation. The dimension of language representation is 768 as in XLM-R base model each token is represented by a 768-dimension vector. We reduce the 768-dimension vectors $v_{1:108}$ to 2-dimension (denoted as $\tilde{v}_{1:108}$) for visualization with the t-SNE algorithm implemented in Scikit-learn Python package \footnote{\url{https://scikit-learn.org/stable/}} with default parameters. We use min-max normalization to normalize $\tilde{v}_{1:108}$ to range $[0,1]$. The 2-dimension language representation are visualized in Figure~\ref{Fig.main_vis}.
\begin{figure*}
\centering  
\subfigure[(Subject, Object, Verbal) Order]{
\label{Fig.sub.1}
\includegraphics[width=0.3\textwidth]{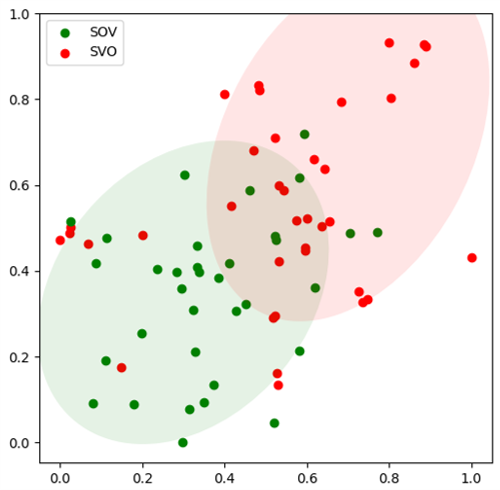}}
\subfigure[Adjective Position]{
\label{Fig.sub.2}
\includegraphics[width=0.3\textwidth]{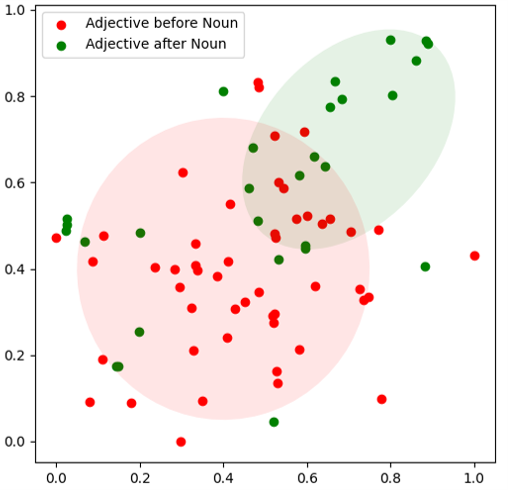}}
\subfigure[Apposition Position]{
\label{Fig.sub.3}
\includegraphics[width=0.3\textwidth]{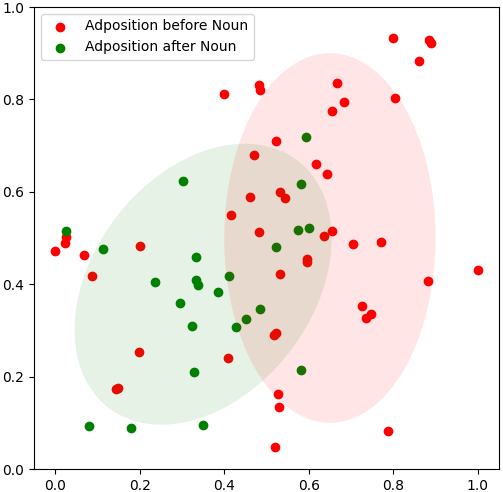}}
\caption{Visualization of the relationship of our language representation and 3 language syntactic features. languages with the same syntactic features approximately fall in the same region. i.e., have similar language representation. Note that syntactic feature data of several languages is not available. Best viewed in color.}
\label{Fig.syn}
\end{figure*}
\subsection{Linguistics Analysis}

 We find that our representation can reflect linguistic similarity and relatedness between languages from different perspectives. We link language representation with several linguistic language similarity measures, also with some linguistic clustering methods. We believe that representation sprachbund is a new paradigm for clustering similar and related languages in linguistics. 
In Figure~\ref{Fig.main_vis}, each language corresponds to one point (2-dimension vector) on that figure. We label all points with the ISO 639-1 code\footnote{\url{http://www.infoterm.info/standardization/ISO_639.php}} of their corresponding languages. 

\paragraph{Relationship with Language Family}
We find that the distribution of language representation has great overlap and similarity with language family.
In Figure~\ref{Fig.main_vis}, the color of each point indicates the language family of its corresponding language. 108 languages are categorized into 22 language families according to Ethnologue\footnote{\url{https://www.ethnologue.com/browse/families}}. All 108 languages and their corresponding language family can be found in Appendix \ref{sec:a}. We only label 10 language families with ellipses for clarity.
Though there are some special cases, languages within the same language family approximately fall in the same region.
\paragraph{Relationship with Geographical Sprachbund}
Geographical sprachbund is a group of similar languages from geographical proximity and language contact, while our representation sprachbund is a group of similar languages from representation proximity. We find that our language representation distribution is consistent with many geographical sprachbunds. In Figure~\ref{Fig.main_vis}, on the top-right, Romance and Germanic language representation closeness can be linked with the Western Europe sprachbund from \citet{10.2307/42581315}; on the bottom-left, the closeness between Indo-Aryan, Dravidian and some Sino-Tibetan languages aligns with Indian subcontinent sprachbund proposed in \citet{10.2307/410649}; on the middle, the similar representation from Turkic, Uralic and Mongolic (mn) also match Altaic sprachbund by \citet{10.2307/41926545}.
\paragraph{Relationship with Lexicostatistics}
We find that similarity in language representation also reflects lexical similarity between languages. We collect lexical similarity data from Ethnologue on several languages (en, fr, de, pt, ro, ru, es, ca), denoted as $Sim_{\textrm{lex}}$. We denote the similarity quantified by our language representation as $Sim_{\textrm{data}}$. We find that $Sim_{\textrm{lex}}$ and $Sim_{\textrm{data}}$ are strongly linear correlated, with Pearson correlation coefficient of 0.83. Lexicostatistics is a method to measure lexical similarity by comparing the percentage of lexical cognates between languages~\citep{10.2307/2739673}, which is very time-consuming. Our language representation can even further help linguists infer lexical similarity more easily (\emph{e.g.} linear regression between representation similarity and lexical similarity). The similarity data is shown in Appendix \ref{sec:b}.
\paragraph{Relationship with Language Syntax}
Languages have diverse syntactic features defined by linguists and can be classified through these features. We show that the distribution of our language representation implies the syntactic features of corresponding languages. We use the lang2vec Python package~\citep{littell-etal-2017-uriel} to query the URIEL database\footnote{\url{http://www.cs.cmu.edu/~dmortens/uriel.html}}. We choose three representative syntactic features: (subject, object, verbal) word order, adjective position and adposition position. As shown in Figure~\ref{Fig.syn}, we find that languages with the same syntactic features approximately have similar language representation.

\paragraph{Surprise: Help for Exploring Linguistic Mystery}
Coincidentally, we find that our language representation connect with an existing under-explored linguistic mystery. In Figure~\ref{Fig.main_vis}, Uralic and Austronesian languages (jv, id, ms) have similar language representation. To the best of our knowledge, only a few linguistic works~\citep{inproceedings,Ohnishi2009} discussed their similarity and relatedness. The reason for their similarity cannot be explained by language family (genetic relationship) or geographical sprachbund (geographical relationship). Their language representation similarity may be a clue that motivates linguists to find more similarity between them and further explain how their similarity formed. 

With the above linguistic analysis, we show that our language representation contain rich linguistic genealogical, geographical, typological, and lexical features of languages, therefore the similarity between language representation can be a good metric for clustering languages. With a 768-dimension vector as numerical feature for each language, we can implement clustering algorithms to cluster similar languages into a representation sprachbund. Our language representation will be released later.

\section{Representation Sprachbund For Multilingual Pre-training}
\subsection{Datasets}

We collect massive multilingual corpora for pre-training and use four datasets for downstream task evaluation.  The multilingual corpora has been described in Section~\ref{RSBst}. We use XNLI~\citep{Conneau2018XNLIEC}, PAWS-X~\citep{Yang2019paws-x}, NER~\citep{Pan2017}, Part of Speech Tagging (POS), MLQA~\citep{Lewis2019MLQAEC}, TydiQA~\citep{Clark2020tydiqa}, XQuAD\citep{artetxe2020cross} and cross-lingual sentence retrieval~\citep{artetxe2019massively} as downstream tasks. For cross-lingual sentence retrieval, we collect 21 language pairs and extract 1000 sentence-pairs for each language-pair from tatoeba\footnote{\url{https://tatoeba.org/eng/downloads}}. This task aims to find the nearest neighbor for each sentence in the other language. 
\begin{table}[H]
\centering
\small
\setlength{\tabcolsep}{2.5pt}{\begin{tabular}{lll}
\hline
\multicolumn{3}{l}{\#$i$ is the $i$th representation sprachbund}                                                                                                                                                 \\ \hline
\#1                   & \multicolumn{2}{l}{\begin{tabular}[c]{@{}l@{}}af als an ast bar br ca ceb da de en eo es el fr fy ga gd \\gl  ia it ku lb nds nl nn no oc pt ro scn sco sq sv tl ur war\end{tabular}} \\ \hline
\#2                   & \multicolumn{2}{l}{\begin{tabular}[c]{@{}l@{}}ar arz bg bs cy fa hi hr id is mg mk ms ps ru sh sl so \\sr su sw yi\end{tabular}}                                                     \\ \hline
\#3                   & \multicolumn{2}{l}{\begin{tabular}[c]{@{}l@{}}am as be ckb cs et eu fi he hu ja jv km la lo lt lv mr my \\ ne or pa pl sa sd sk th uk wuu zh\end{tabular}}                            \\ \hline
\#4                   & \multicolumn{2}{l}{az bn gu hy ka kk kn ko ky ml mn si ta te tt ug uz vi tr}                                                                                                          \\ \hline
\end{tabular}}
\caption{Components of 4 representation sprachbunds}
\label{RSBmem}
\end{table}

\begin{table*}[]
\centering
\setlength{\tabcolsep}{3.5pt}{\begin{tabular}{cccccccc}
\hline
Tasks    & XNLI(Acc)          & PAWS-X(Acc)       & POS(F1)       & NER(F1)          & TydiQA(F1)        & XQuAD(F1)        & MLQA(F1)         \\ \hline
XLM-R    & 75.0          & 84.2          & 79.7          & 60.8          & 45.9          & 70.3          & 65.0          \\
XLM-R CT & 74.8          & 84.8          & 79.9          & 60.9          & 46.3          & 70.5          & 65.4          \\
Random   & 74.8          & 84.8          & 79.4          & 60.9          & 47.2          & 70.2          & 65.5          \\
Ours     & \textbf{75.7} & \textbf{85.4} & \textbf{80.1} & \textbf{63.5} & \textbf{47.8} & \textbf{70.9} & \textbf{66.3} \\ \hline
\end{tabular}}
\caption{Performance of our model fine-tune with English on 7 cross-lingual understanding tasks. XLM-R: directly fine-tune on XLM-R model. XLM-R CT: continue to pre-train XLM-R base model with our corpora. Random: pre-train one model for each random language cluster. Ours: pre-train one model for each representation sprachbund respectively.}
\label{all}
\end{table*}
\begin{table*}[]
\scriptsize
\centering
\setlength{\tabcolsep}{2.0pt}{%
\begin{tabular}{l|llllllllllllllllll|l|ll|l}
\hline
Languages & de-en & pt-en & es-en & nl-en & tl-en & ur-en & el-en & af-en & fr-en & it-en & nl-de & de-it & es-pt & de-el & sv-da & da-no & fr-de & it-ro & ar-ru & zh-ja & pl-cs & Avg  \\ \hline
XLM-R     & 89.0  & 78.7  & 72.0  & 77.4  & 29.9  & 33.4  & 54.1  & 53.1  & 73.2  & 65.9  & 67.0  & 55.1  & 75.9  & 54.5  & 81.1  & 89.6  & 76.0  & 51.2  & 43.4  & 52.6  & 73.8  & 64.1 \\
XLM-R CT  & 85.9  & 75.5  & 70.2  & 77.3  & 31.4  & 36.5  & 52.4  & 57.9  & 73.6  & 63.5  & 65.4  & 54.2  & 73.2  & 49.9  & 79.6  & 89.6  & 72.3  & 50.4  & 46.7  & 59.6  & 76.8  & 63.9 \\
Ours      & 90.3  & 80.6  & 75.3  & 79.0  & 32.2  & 39.8  & 54.5  & 57.4  & 74.6  & 67.9  & 68.2  & 57.8  & 76.7  & 55.3  & 81.0  & 89.9  & 77.5  & 53.3  & 58.7  & 61.7  & 82.1  & \textbf{67.3} \\ \hline
\end{tabular}
}
\caption{Performance (Accuracy) of our model on cross-lingual sentence retrieval task without fine-tuning. Languages in different representation sprachbunds are separated with vertical lines.}
\label{tatoeba}
\end{table*}

\subsection{Settings}
We use the XLM-R base model as our base model. Fairseq\footnote{\url{https://github.com/pytorch/fairseq}} is used as our pre-training code base. The Huggingface Transformers\footnote{\url{https://huggingface.co/transformers/}} is used as our fine-tuning code base.
We use the hierarchical clustering algorithm implemented by Scikit-learn Python package\footnote{\url{https://scikit-learn.org/stable/}} for clustering language representation. We cluster languages into 4 representation sprachbunds. The reason for clustering 4 representation sprachbunds is shown in Section~\ref{reason}. Components of representation sprachbunds are shown in Table~\ref{RSBmem}.
We use the shared vocabulary of XLM-R base model for reusing the pre-trained parameters and keeping the comparability with the baseline model. 

\paragraph{Pre-training Setting} We initialize our model with the XLM-R base model parameters and run continual pre-training for 40000 updates on 8 Nvidia V100 GPUs with total batch size 8192. The experiment takes about 8 days. We use Adam optimizer with a linear warm-up and set the learning rate to 3e-5. We pre-train 4 models according to our 4 representation sprachbund corpora. We also randomly create 4 language clusters (each with the same language number  as the representation sprachbund). We pre-train 4 models with random language clusters as baseline. To evaluate the impact of continual pre-train corpora, we also continue to pre-train XLM-R base model with our corpora.

\paragraph{Downstream Task Setting} 
For XNLI, we set the learning rate to 5e-6 and train 10 epochs with batch size 32. For POS tagging, we set the learning rate to 2e-5 and train 20 epochs with batch size 32. For MLQA, we set the learning rate to 3e-5, train 2 epochs following BERT for SQuAD with batch size 12. For PAWS-X, NER, TydiQA and XQuAD, we follow the default settings in XTREME~\citep{hu2020xtreme}.The downstream task performance on one specific language is measured by fine-tuning the model pre-trained with the corresponding representation sprachbund corpora. We select the checkpoint with the best performance on the dev set. The results are averaged over three runs. For cross-lingual sentence retrieval, we use the cosine similarity of the average middle layer (the 7th layer of our 12-layer model) embedding for retrieval without fine-tuning. 

\begin{table*}[]

\small
\centering

\setlength{\tabcolsep}{4.5pt}{\begin{tabular}{ccccccccccccccccc}
\hline
\multicolumn{1}{l|}{Languages} & en                       & fr                       & es                       & \textbf{de}              & el   & \multicolumn{1}{c|}{ur}   & bg   & \textbf{ru} & ar   & hi   & \multicolumn{1}{c|}{sw}   & tr & \multicolumn{1}{c|}{\textbf{vi}}  & \textbf{th} & \multicolumn{1}{c|}{zh}   & Avg           \\ \hline
\multicolumn{17}{c}{XNLI fine-tune with English}                                                                                                                                                                                                                                                                                                 \\ \hline
\multicolumn{1}{c|}{XLM-R}     & 84.8                     & 78.9                     & 79.2                     & 77.7                     & 76.6 & \multicolumn{1}{c|}{67.0} & 78.6 & 76.6        & 72.3 & 71.1 & \multicolumn{1}{c|}{66.4} & 73.3        & \multicolumn{1}{c|}{75.6} & 72.4        & \multicolumn{1}{c|}{74.7} & 75.0          \\
\multicolumn{1}{l|}{XLM-R CT}  & \multicolumn{1}{l}{84.5} & \multicolumn{1}{l}{78.4} & \multicolumn{1}{l}{79.5} & \multicolumn{1}{l}{77.2} & 76.4 & \multicolumn{1}{c|}{66.8} & 78.3 & 76.2        & 73.0 & 70.6 & \multicolumn{1}{c|}{65.4} & 73.3        & \multicolumn{1}{c|}{75.3} & 72.6        & \multicolumn{1}{c|}{74.2} & 74.8          \\
\multicolumn{1}{c|}{Random}    & 84.3                     & 78.0                     & 78.5                     & 77.1                     & 76.5 & \multicolumn{1}{c|}{67.3} & 78.2 & 76.2        & 72.2 & 70.5 & \multicolumn{1}{c|}{66.4} & 73.5        & \multicolumn{1}{c|}{75.4} & 72.8        & \multicolumn{1}{c|}{74.7} & 74.8          \\
\multicolumn{1}{c|}{Ours}      & \multicolumn{1}{l}{84.1} & \multicolumn{1}{l}{78.4} & \multicolumn{1}{l}{80.0} & \multicolumn{1}{l}{77.8} & 77.3 & \multicolumn{1}{c|}{68.2} & 78.4 & 76.7        & 74.5 & 72.1 & \multicolumn{1}{c|}{68.6} & 74.9        & \multicolumn{1}{c|}{75.3} & 74.1        & \multicolumn{1}{c|}{74.6} & \textbf{75.7} \\ \hline
\multicolumn{17}{c}{XNLI fine-tune with pivot language}                                                                                                                                                                                                                                                                                          \\ \hline
\multicolumn{1}{c|}{XLM-R}     & 83.6                     & 79.5                     & 80.0                     & 79.8                     & 77.4 & \multicolumn{1}{c|}{69.3} & 80.1 & 78.4        & 74.2 & 73.3 & \multicolumn{1}{c|}{66.9} & 74.2        & \multicolumn{1}{c|}{78.8} & 76.1        & \multicolumn{1}{c|}{76.3} & 76.5          \\
\multicolumn{1}{c|}{Ours}      & \multicolumn{1}{l}{84.2} & \multicolumn{1}{l}{80.0} & \multicolumn{1}{l}{80.4} & \multicolumn{1}{l}{80.4} & 77.8 & \multicolumn{1}{c|}{69.9} & 80.2 & 78.0        & 75.5 & 73.3 & \multicolumn{1}{c|}{70.9} & 75.0        & \multicolumn{1}{c|}{78.3} & 76.4        & \multicolumn{1}{c|}{75.7} & \textbf{77.1} \\ \hline
\multicolumn{17}{c}{XNLI fine-tune with every language}                                                                                                                                                                                                                                                                                          \\ \hline
\multicolumn{1}{c|}{XLM-R}     & 84.6                     & 79.5                     & 80.9                     & 79.8                     & 79.1 & \multicolumn{1}{c|}{67.1} & 80.2 & 78.4        & 75.5 & 73.9 & \multicolumn{1}{c|}{71.1} & 77.2        & \multicolumn{1}{c|}{78.8} & 76.1        & \multicolumn{1}{c|}{78.4} & 77.4          \\
\multicolumn{1}{c|}{Ours}      & \multicolumn{1}{l}{84.1} & \multicolumn{1}{l}{80.4} & \multicolumn{1}{l}{81.1} & \multicolumn{1}{l}{80.4} & 78.9 & \multicolumn{1}{c|}{66.5} & 80.3 & 78.0        & 77.2 & 74.3 & \multicolumn{1}{c|}{72.6} & 76.9        & \multicolumn{1}{c|}{78.3} & 76.4        & \multicolumn{1}{c|}{78.0} & \textbf{77.6} \\ \hline
\end{tabular}}
\caption{Performance (Accuracy) of our model on XNLI dataset on three settings. Note that we do not use the result in~\citep{conneau-etal-2020-unsupervised}, instead we fine-tune the model in our settings. Languages in different representation sprachbunds are separated with vertical lines. Pivot languages are \textbf{bold} in the first row.}
\label{table2}
\end{table*}

\paragraph{Evaluation Setting} There are three main settings in the fine-tuning stage. (i) \textbf{Fine-tune with English.} We fine-tune the model with English labeled data of downstream task. (ii) \textbf{Fine-tune with every language.} \emph{e.g.} We fine-tune the model with the French labeled data and test its performance on the French test set. (iii) \textbf{Fine-tune with pivot language.} We choose a pivot language $l_i$ in its representation sprachbund $m$ based on similarity with other languages (if $i=\mathop{\arg\max}_{i} \sum_{l_j\in L_m} cos(v_i,v_j)$), and fine-tune the model with the pivot language labeled data.
\subsection{Main Results}



We conduct experiments in several settings on different types of tasks. We find that our approach obtains significant improvements over XLM-R base model and randomly clustered model when applied in pre-training and downstream tasks.

\paragraph{Improving Pre-Training}

\begin{table*}[hbt]
\centering
\small

\setlength{\tabcolsep}{4.5pt}{\begin{tabular}{c|cccccc|ccccc|cc|cc|c}
\hline
Languages & en   & fr   & es   & de   & el   & ur   & bg   & ru   & ar   & hi   & sw   & tr   & vi   & th   & zh   & Avg           \\ \hline
XLM-R+All & 84.8 & 81.3 & 82.0 & 80.5 & 80.0 & 71.7 & 81.6 & 79.1 & 78.2 & 75.6 & 73.1 & 78.1 & 79.4 & 77.2 & 79.8 & 78.8          \\
XLM-R+RSB & 85.3 & 80.9 & 81.8 & 80.6 & 80.0 & 71.2 & 81.3 & 79.4 & 77.7 & 75.9 & 72.6 & 77.5 & 79.0 & 77.4 & 78.3 & 78.6          \\
RSB+      & 85.6 & 81.4 & 81.7 & 80.5 & 80.9 & 70.9 & 82.0 & 79.5 & 78.0 & 76.8 & 73.8 & 77.6 & 78.7 & 77.3 & 79.7 & \textbf{79.0} \\ \hline
\end{tabular}}
\caption{Downstream task (XNLI) data efficiency of our model. XLM-R+All: use XNLI data in all languages to fine-tune XLM-R base model. XLM-R+RSB: use XNLI data in the same representation sprachbund to fine-tune XLM-R base model. RSB+: use XNLI data in the same representation sprachbund to fine-tune the pre-trained model. Different representation sprachbunds are separated with vertical lines. }
\label{table5}
\end{table*}
In Table~\ref{all}, We find that our approach significantly outperforms all baseline  models on 7 cross-lingual tasks. The 7 tasks are representative of almost all kinds of cross-lingual understanding tasks, which shows the universal effectiveness of our model. The detailed results of each task and each language are shown in Appendix \ref{sec:c}.
\paragraph{Improving Fine-Tuning}
In Table~\ref{table2}, we also show that without additional costly continual pre-training, directly fine-tuning multiple models with representation sprachbund also improves the performance. We fine-tune each model (XLM-R base) with the labeled pivot language data on XNLI of each representation sprachbund. In Table~\ref{table2}, we find that the improvement is significant (from 75.0 to 76.5).
  We also show
that fine-tuning the model with the labeled data from each
language yields significantly better results (77.4). We conclude that fine-tuning with the language similar to target language is likely to boost the performance.

\paragraph{Improving Multilingual Embeddings} In Table~\ref{tatoeba}, we show that the performance on cross-lingual sentence retrieval greatly improves with our approach. Though our method is not designed for improving multilingual embeddings, better multilingual embeddings are generated without additional fine-tuning.

\begin{table}[H]
\centering
\small

\setlength{\tabcolsep}{2pt}{
\begin{tabular}{cccccccc}
\hline
Task & XNLI         & PAWS-X & POS          & NER          & TydiQA       & XQuAD        & MLQA         \\ \hline
High & 0.2          & 0.6    & 0            & 1.3          & \textbf{1.9} & 0.5          & 1.0          \\
Low  & \textbf{1.2} & 0.6    & \textbf{1.2} & \textbf{3.1} & 1.4          & \textbf{0.8} & \textbf{1.5} \\ \hline
\end{tabular}}

\caption{Low resource languages has more significant gains with our approach. Low indicates the languages are low resource and isolated.}
\label{lowrc}
\end{table}
\subsection{Analysis}
\paragraph{Gaining More on Low Resource Languages} 
In Table~\ref{lowrc}, we show that our approach brings more gains to those low resource and isolated languages (including ur, ar, sw, tr, vi, th, zh, pl, ja, ko, id, fi, bn, te, tl, af, ms, fa, mr, et, he, jv, eu, yo, my, hu, ta, ml, kk ,kn, ka) compared with those high resource languages (including en, de, es, ru, bg, hi, it, fr, nl, pt) which will be beneficial for bridging the large performance gap between high resource and low resource languages. An intuitive explanation is that those low resource and isolated languages suffer from more serious cross-lingual contradictions when trained with those dissimilar high resource languages. When clustered with similar languages, those low resource and isolated languages are likely to benefit a lot.

\paragraph{Achieving Data Efficiency in Downstream Tasks} We show that with the continual pre-training step with representation sprachbund corpora, less data for downstream tasks is needed to achieve high accuracy.
In Table~\ref{table5}, on XNLI task, we find that fine-tuning our pre-trained model with all the downstream task data from each representation sprachbund (less than 30\% of all the data) achieves better results than fine-tuning XLM-R with all the data (from 78.8 to 79.0).
We also find that fine-tuning XLM-R with the representation sprachbund data achieves results comparable with fine-tuning using all the data (78.6 and 78.8), which means that using data from similar languages (though less) works well.



\begin{figure}[H] 
\centering 
\includegraphics[width=0.45\textwidth]{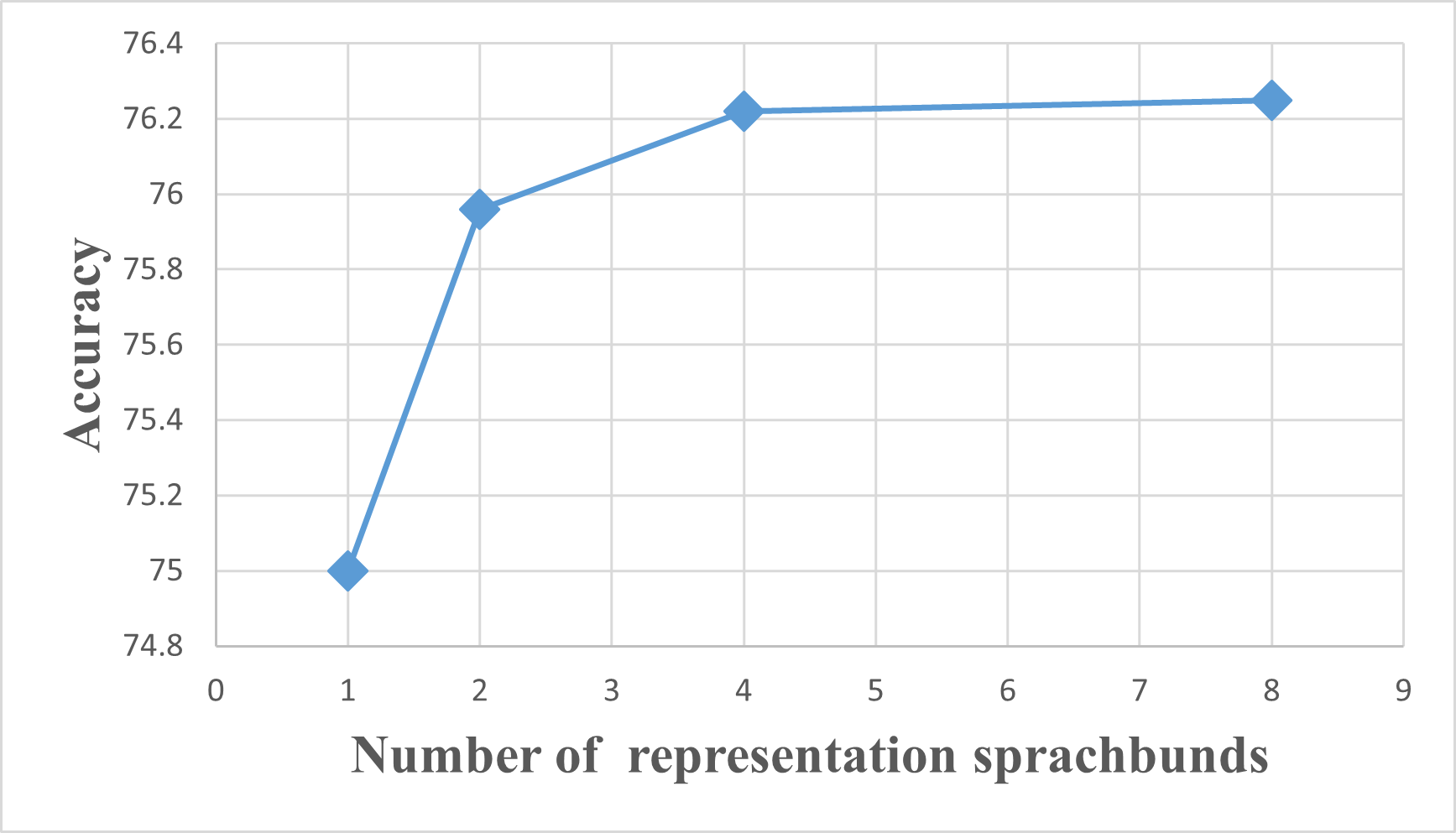} 
\caption{Impact of the number of representation sprachbunds on the performance. Performance (Accuracy) is measured by fine-tuning with English labeled data on XNLI task.} 
\label{Fig.numb} 
\end{figure}

\paragraph{Choosing the Number of Representation Sprachbunds}
\label{reason}
We conduct experiments to choose the number of representation sprachbunds. We cluster languages into 1,2,4,8 representation sprachbunds, and pre-train 1,2,4,8 models, respectively. We evaluate through fine-tuning with English on the XNLI dataset. As shown in Figure~\ref{Fig.numb}, with the increase of the number of representation sprachbunds, the performance also increases. We find that clustering languages into 4 representation sprachbunds is a desirable choice, as from 4 to 8 little gain is obtained but the cost doubles.

\section{Conclusion}To reduce the cross-lingual contradictions in pre-training one model for all languages, we propose to merge similar languages into a representation sprachbund and pre-train one model for each representation sprachbund. Results show that our approach outperforms strong baselines in various settings and tasks. We also identify the relationship between our representation sprachbund with linguistic theories. Applications of our representation sprachbund as a paradigm for clustering languages in linguistics will be explored in subsequent work.

\section*{Acknowledgements}
We thank Ning Wu, Xiaoze Jiang, Yuan Chai, Junhe Zhao, Shunyu Zhang for their useful suggestions on writing this paper.
\bibliography{anthology,custom}
\bibliographystyle{acl_natbib}

\appendix
\clearpage
\section{Linguistic Language Family of All Languages}
\label{sec:a}

\begin{table}[H]
\centering
\begin{tabular}{c|c}
\hline
Language Family             & Languages                                                                                                             \\ \hline
{Germantic}  & {\begin{tabular}[c]{@{}c@{}}af,als,bar,cy,da,de,en,fy,gd,is,\\ lb,nds,nl,nn,no,sco,sv,yi\end{tabular}} \\
                            &                                                                                                                       \\
Greek                       & el                                                                                                                    \\
Japonic                     & ja                                                                                                                    \\
Sino-Tibetan                & my,wuu,zh                                                                                                             \\
Turkic                      & az,kk,ky,tr,tt,ug,uz                                                                                                  \\
Uralic                      & et,fi,hu                                                                                                              \\
Austroasiatic               & km,vi,war                                                                                                             \\
Dravidian                   & kn,ml,ta,te                                                                                                           \\
{Slavic}     & {\begin{tabular}[c]{@{}c@{}}be,bg,bs,cs,hr,lt,lv,mk,\\ pl,ru,sh,sk,sl,sr,uk\end{tabular}}              \\
                            &                                                                                                                       \\
Kartvelian                  & ka                                                                                                                    \\
Niger-Congo                 & sw                                                                                                                    \\
Austronesian                & ceb,id,jv,mg,ms,su,tl                                                                                                 \\
Armenian                    & hy                                                                                                                    \\
Koreanic                    & ko                                                                                                                    \\
Albanian                    & sq                                                                                                                    \\
Tai-Kadai                   & lo,th                                                                                                                 \\
{Romance}    & {\begin{tabular}[c]{@{}c@{}}an,ast,br,ca,es,fr,gl,\\ it,la,oc,pt,ro,scn,eu\end{tabular}}               \\
                            &                                                                                                                       \\
Constructed                 & eo,ia                                                                                                                 \\
Afro-Asiatic                & am,ar,arz,he,so                                                                                                       \\
Celtic                      & ga                                                                                                                    \\
{Indo-Aryan} & {\begin{tabular}[c]{@{}c@{}}as,bn,ckb,fa,gu,hi,\\ ku,mr,ne,or,pa,ps\end{tabular}}                      \\
                            &                                                                                                                       \\
Mongolic                    & sa,sd,si,ur                                                                                                           \\ \hline
\end{tabular}
\caption{Languages and their corresponding language family}

\end{table}
\section{Lexical similarity and Embedding similarity}
\label{sec:b}

\begin{table}[H]
\small
\centering
\begin{tabular}{ccccccccc}
\hline
Languages & ca   & en   & fr   & de   & pt   & ro   & ru   & es   \\ \hline
ca        & 1.00 & -    & 0.85 & -    & 0.85 & 0.73 & -    & 0.85 \\
en        & -    & 1.00 & 0.27 & 0.60 & -    & -    & 0.24 & -    \\
fr        & 0.85 & 0.27 & 1.00 & 0.28 & 0.75 & 0.75 & -    & 0.75 \\
de        & -    & 0.60 & 0.28 & 1.00 & -    & -    & -    & -    \\
pt        & 0.85 & -    & 0.75 & -    & 1.00 & 0.72 & -    & 0.88 \\
ro        & 0.73 & -    & 0.75 & -    & 0.72 & 1.00 & 0.72 & 0.71 \\
ru        & -    & 0.24 & -    & -    & -    & -    & 1.00 & -    \\
es        & 0.85 & -    & 0.75 & -    & 0.88 & 0.71 & -    & 1.00 \\ \hline
\end{tabular}
\caption{Lexical similarity from Ethnologue. Some data is missing.}

\end{table}
\begin{table}[H]
\small
\centering

\begin{tabular}{cllllllll}
\hline
Languages & \multicolumn{1}{c}{ca} & \multicolumn{1}{c}{en} & \multicolumn{1}{c}{fr} & \multicolumn{1}{c}{de} & \multicolumn{1}{c}{pt} & \multicolumn{1}{c}{ro} & \multicolumn{1}{c}{ru} & \multicolumn{1}{c}{es} \\ \hline
ca        & 1.00                   & 0.08                   & 0.76                   & 0.22                   & 0.63                   & 0.56                   & 0.23                   & 0.81                   \\
en        & 0.08                   & 1.00                   & 0.26                   & 0.38                   & 0.28                   & 0.17                   & 0.31                   & 0.00                   \\
fr        & 0.76                   & 0.26                   & 1.00                   & 0.49                   & 0.63                   & 0.65                   & 0.47                   & 0.68                   \\
de        & 0.22                   & 0.38                   & 0.49                   & 1.00                   & 0.47                   & 0.49                   & 0.59                   & 0.26                   \\
pt        & 0.63                   & 0.28                   & 0.63                   & 0.47                   & 1.00                   & 0.61                   & 0.45                   & 0.64                   \\
ro        & 0.56                   & 0.17                   & 0.65                   & 0.49                   & 0.61                   & 1.00                   & 0.48                   & 0.56                   \\
ru        & 0.23                   & 0.31                   & 0.47                   & 0.59                   & 0.45                   & 0.48                   & 1.00                   & 0.24                   \\
es        & 0.81                   & 0.00                   & 0.68                   & 0.26                   & 0.64                   & 0.56                   & 0.24                   & 1.00                   \\ \hline
\end{tabular}
\caption{Language embedding similarity}
\end{table}

\section{Detailed Results of Cross-lingual Tasks}
\label{sec:c}

\begin{table}[H]

\centering
\setlength{\tabcolsep}{2.5pt}{\begin{tabular}{c|cc|ccc|c|c|c}
\hline
Languages & ar   & hi   & de   & en   & es   & zh   & vi   & Avg           \\ \hline
XLM-R     & 55.3 & 61.3 & 62.1 & 80.0 & 68.1 & 61.5 & 66.9 & 65.0          \\
XLM-R CT  & 57.1 & 61.7 & 61.9 & 80.1 & 68.0 & 61.5 & 67.3 & 65.4          \\
Random    & 57.2 & 62.3 & 62.0 & 81.0 & 67.6 & 61.6 & 67.2 & 65.5          \\
Ours      & 59.2 & 63.1 & 62.9 & 80.4 & 69.3 & 61.4 & 67.6 & \textbf{66.3} \\ \hline
\end{tabular}}
\caption{Performance (F1 score) on MLQA dataset. We examine the results on MLQA on fine-tuning with English setting. Languages in different embedding sprachbunds are separated with vertical lines. }
\label{mlqa}
\end{table}
\begin{table}[H]
\small
\centering
\begin{tabular}{c|cccc|cc|c|c}
\hline
Languages & de   & en   & es   & fr   & ja   & zh   & ko   & Avg           \\ \hline
XLM-R     & 87.5 & 94.4 & 88.5 & 88.5 & 75.9 & 80.1 & 74.7 & 84.2          \\
XLM-R CT  & 87.7 & 94.4 & 88.8 & 88.5 & 77.2 & 81.1 & 75.6 & 84.8          \\
Random    & 87.0 & 94.4 & 88.8 & 88.9 & 77.4 & 80.0 & 76.7 & 84.8          \\
Ours      & 88.4 & 94.9 & 89.1 & 89.6 & 78.2 & 80.1 & 77.2 & \textbf{85.4} \\ \hline
\end{tabular}
\caption{Performance (F1 score) on PAWS-X dataset. We examine the results on PAWS-X on fine-tuning with English setting. Languages in different embedding sprachbunds are separated with vertical lines. }
\end{table}
\begin{table*}
\small
\centering
\setlength{\tabcolsep}{2.0pt}{\begin{tabular}{c|llll|lllllllll|lll|ll|l}
\hline
Languages                     & \multicolumn{1}{c}{ar}   & \multicolumn{1}{c}{ru}   & \multicolumn{1}{c}{bg}   & \multicolumn{1}{c|}{hi}   & \multicolumn{1}{c}{ur}   & \multicolumn{1}{c}{de}   & \multicolumn{1}{c}{el}   & \multicolumn{1}{c}{en}   & \multicolumn{1}{c}{es}   & \multicolumn{1}{c}{fr}   & \multicolumn{1}{c}{it}   & \multicolumn{1}{c}{nl}   & \multicolumn{1}{c|}{pt}   & \multicolumn{1}{c}{th}   & \multicolumn{1}{c}{zh}   & \multicolumn{1}{c|}{pl}   & \multicolumn{1}{c}{tr}   & \multicolumn{1}{c|}{vi}   & \multicolumn{1}{c}{Avg}  \\ \hline
XLM-R                         & \multicolumn{1}{c}{69.5} & \multicolumn{1}{c}{86.6} & \multicolumn{1}{c}{88.5} & \multicolumn{1}{c|}{72.2} & \multicolumn{1}{c}{59.8} & \multicolumn{1}{c}{92.3} & \multicolumn{1}{c}{87.6} & \multicolumn{1}{c}{96.4} & \multicolumn{1}{c}{88.8} & \multicolumn{1}{c}{89.3} & \multicolumn{1}{c}{92.1} & \multicolumn{1}{c}{88.6} & \multicolumn{1}{c|}{90.0} & \multicolumn{1}{c}{58.6} & \multicolumn{1}{c}{59.7} & \multicolumn{1}{c|}{84.0} & \multicolumn{1}{c}{73.6} & \multicolumn{1}{c|}{56.7} & \multicolumn{1}{c}{79.7} \\
\multicolumn{1}{l|}{XLM-R CT} & 69.4                     & 86.1                     & 88.6                     & 68.2                      & 61.5                     & 91.5                     & 88.5                     & 96.4                     & 89.6                     & 89.8                     & 92.2                     & 88.9                     & 90.0                      & 57.8                     & 62.9                     & 84.0                      & 74.1                     & 57.9                      & 79.9                     \\
Random                        & 68.5                     & 86.5                     & 88.6                     & 69.2                      & 59.7                     & 92.3                     & 87.9                     & 96.3                     & 88.1                     & 88.9                     & 92.0                     & 88.4                     & 90.0                      & 58.2                     & 60.5                     & 83.9                      & 72.9                     & 57.1                      & 79.4                     \\
Ours                          & 69.3                     & 85.4                     & 88.2                     & 70.8                      & 60.1                     & 92.4                     & 88.6                     & 95.6                     & 89.1                     & 89.4                     & 92.3                     & 88.9                     & 90.9                      & 58.5                     & 63.0                     & 84.7                      & 75.5                     & 58.5                      & \textbf{80.1}            \\ \hline
\end{tabular}}
\caption{Performance (F1 score) on POS tagging dataset. We examine the results on POS tagging on fine-tuning with English setting. Languages in different embedding sprachbunds are separated with vertical lines. }
\label{table0}
\end{table*}

\begin{table*}
\small
\centering
\begin{tabular}{c|c|cccc|c|ccc|c}
\hline
Languages & en   & ar   & ru   & sw   & id   & fi   & bn   & ko   & te   & Avg           \\ \hline
XLM-R     & 60.7 & 52.5 & 50.1 & 50.2 & 63.8 & 51.2 & 31.1 & 23.0 & 30.9 & 45.9          \\
XLM-R CT  & 60.1 & 53.7 & 50.6 & 54.5 & 66.1 & 52.4 & 31.0 & 21.6 & 26.7 & 46.3          \\
Random    & 61.2 & 52.4 & 50.1 & 51.1 & 65.5 & 50.1 & 38.3 & 22.2 & 33.5 & 47.2          \\
Ours      & 62.8 & 54.6 & 51.6 & 49.3 & 66.1 & 50.8 & 39.1 & 22.6 & 33.7 & \textbf{47.8} \\ \hline
\end{tabular}
\caption{Performance (F1 score) on TydiQA dataset. We examine the results on TydiQA on fine-tuning with English setting. Languages in different embedding sprachbunds are separated with vertical lines. }
\end{table*}
\begin{table*}
\centering
\small
\begin{tabular}{c|cccc|ccc|cc|cc|c}
\hline
Languages & en   & es   & de   & el   & ru   & hi   & ar   & th   & zh   & tr   & vi   & Avg           \\ \hline
XLM-R     & 83.0 & 76.1 & 73.2 & 72.4 & 73.3 & 68.0 & 66.0 & 68.0 & 51.7 & 67.4 & 73.8 & 70.3          \\
XLM-R CT  & 83.4 & 75.9 & 73.4 & 72.4 & 73.7 & 69.2 & 65.8 & 68.1 & 52.8 & 66.7 & 74.0 & 70.5          \\
Random    & 82.8 & 76.0 & 72.8 & 71.9 & 73.2 & 68.9 & 65.5 & 67.4 & 52.6 & 66.8 & 74.1 & 70.2          \\
Ours      & 82.8 & 75.5 & 74.5 & 73.0 & 73.4 & 69.5 & 67.3 & 68.0 & 53.1 & 68.5 & 73.8 & \textbf{70.9} \\ \hline
\end{tabular}
\caption{Performance (F1 score) on XQuAD dataset. We examine the results on XQuAD on fine-tuning with English setting. Languages in different embedding sprachbunds are separated with vertical lines. }
\end{table*}

\begin{table*}
\scriptsize
\centering
\setlength{\tabcolsep}{2.0pt}{\begin{tabular}{c|cccccccccccccccccccc}
\hline
Languages & en            & de            & el            & tl            & af            & nl            & ur            & fr            & pt            & es                                 & \multicolumn{1}{c|}{it}            & ar            & id            & ms            & hi            & fa            & bg            & ru            & sw            & fa            \\ \hline
XLM-R     & 83.0          & 74.3          & 72.5          & 71.4          & 74.6          & 80.4          & 50.1          & 76.9          & 77.9          & 70.8                               & \multicolumn{1}{c|}{77.2}          & 45.1          & 50.1          & 56.4          & 66.1          & 40.7          & 77.3          & 63.6          & 66.9          & 40.7          \\
XLM-R CT  & 82.5          & 74.2          & 72.8          & 70.9          & 74.8          & 80.0          & 53.3          & 77.0          & 77.2          & 71.3                               & \multicolumn{1}{c|}{77.4}          & 47.5          & 47.4          & 62.2          & 65.6          & 42.2          & 76.6          & 63.7          & 65.9          & 42.2          \\
Random    & 82.7          & 74.2          & 72.6          & 71.2          & 74.7          & 80.2          & 51.7          & 77.0          & 77.6          & 71.0                               & \multicolumn{1}{c|}{77.3}          & 46.3          & 48.7          & 59.3          & 65.9          & 41.4          & 77.0          & 63.7          & 66.4          & 41.4          \\
Ours      & \textbf{83.1} & \textbf{76.4} & \textbf{74.2} & \textbf{73.1} & \textbf{78.2} & \textbf{81.8} & 47.2          & \textbf{78.7} & \textbf{78.2} & \textbf{74.5}                      & \multicolumn{1}{c|}{\textbf{78.4}} & \textbf{51.1} & 49.6          & \textbf{68.7} & 65.2          & \textbf{45.4} & \textbf{78.0} & \textbf{64.1} & \textbf{67.0} & \textbf{45.4} \\ \hline
Languages & mr            & et            & ja            & zh            & he            & jv            & eu            & fi            & yo            & \multicolumn{1}{c|}{my}            & hu                                 & ta            & te            & vi            & ml            & tr            & ko            & kk            & bn            & ka            \\ \cline{2-21} 
XLM-R     & 59.5          & 72.3          & 18.3          & 24.7          & 52.1          & 58.5          & 60.2          & 75.3          & 41.5          & \multicolumn{1}{c|}{51.7}          & 76.1                               & 53.7          & 47.0          & 65.6          & 61.3          & 73.9          & 49.7          & 44.6          & 66.3          & 65.5          \\
XLM-R CT  & 60.9          & 71.7          & 18.1          & 21.9          & 50.8          & 58.2          & 60.1          & 74.5          & 38.3          & \multicolumn{1}{c|}{51.2}          & 75.2                               & 52.9          & 48.2          & 65.7          & 60.7          & 74.5          & 49.0          & 49.6          & 66.2          & 64.7          \\
Random    & 60.2          & 72.0          & 18.2          & 23.3          & 51.5          & 58.3          & 60.2          & 74.9          & 39.9          & \multicolumn{1}{c|}{51.4}          & 75.7                               & 53.3          & 47.6          & 65.7          & 61.0          & 74.2          & 49.4          & 47.1          & 66.3          & 65.1          \\
Ours      & \textbf{62.8} & \textbf{73.8} & \textbf{19.4} & 21.9          & \textbf{53.4} & \textbf{63.2} & \textbf{65.7} & \textbf{76.0} & \textbf{51.6} & \multicolumn{1}{c|}{\textbf{56.0}} & \textbf{76.9}                      & \textbf{55.6} & \textbf{55.0} & \textbf{66.8} & \textbf{65.0} & \textbf{77.0} & \textbf{53.0} & \textbf{50.8} & \textbf{70.5} & \textbf{67.1} \\ \hline
\end{tabular}}
\caption{Performance (F1 score) on NER dataset. We examine the results on NER on fine-tuning with English setting. Languages in different embedding sprachbunds are separated with vertical lines. }
\end{table*}
\end{document}